\documentclass[runningheads]{llncs}
\usepackage[T1]{fontenc}
\usepackage{graphicx}
\usepackage{amsmath}
\usepackage{amssymb}
\usepackage{array}
\usepackage{graphicx}
\usepackage{subcaption}

\begin{document}

\title{Mind the Gap: Quantifying the Domain Gap in Cross-Sensor Diffusion Super-Resolution}

\author{
Dawid Kopeć\inst{1} \orcidID{0009-0000-1765-5810} \and
Katarzyna Jabłońska\inst{1} \orcidID{0009-0006-6824-9437} \and
Wojciech Kozłowski\inst{1} \orcidID{0009-0009-1532-4415} \and \\
Maciej Zięba\inst{1,2} \orcidID{0000-0003-4217-7712}
}

\authorrunning{D. Kopeć et al.}

\institute{
WUST, Wybrzeże Stanisława Wyspiańskiego 27, 50-370 Wrocław, Poland 
\email{\{dawid.kopec, katarzyna.jablonska, wojciech.kozlowski, maciej.zieba\}@pwr.edu.pl}
\and
Tooploox, Tęczowa 7, 53-601 Wrocław, Poland
}

\titlerunning{Quantifying the Domain Gap in Cross-Sensor Diffusion Super-Resolution}

\maketitle

\begin{abstract}
Demand for high-resolution satellite imagery has increased interest in super-resolution (SR) to bridge the spatial resolution gap between freely available missions such as Sentinel-2 and commercial systems like PlanetScope. 
Because no sensor provides true paired low- and high-resolution observations, SR models are usually trained on synthetically degraded data, creating a domain gap on real cross-sensor imagery. In this work, we provide the first systematic study of how this synthetic-to-real mismatch affects the performance of modern diffusion-based SR models. Using a large, geometrically and temporally aligned dataset of Sentinel-2 and PlanetScope imagery, we evaluate five state-of-the-art diffusion architectures under controlled experimental settings. We also introduce \textit{LPIPS\textsubscript{Sat}}, a domain-adapted perceptual metric based on Sentinel-2 self-supervised features. Our results show two persistent challenges: synthetically trained models degrade sharply on real pairs, while models trained on real cross-sensor data exhibit optimisation difficulties and struggle to adapt to the physical and radiometric diversity. These findings highlight a key limitation of current SR and motivate methods that disentangle super-resolution from domain adaptation.

\end{abstract}

\section{Introduction}

Remote sensing (RS) has become critical for decision-making in applications such as wildfire monitoring, mineral exploration, and water resource management~\cite{ma2019deep,shanmugapriya2019applications,sun2024application}. 
Despite the increasing availability of satellite imagery, a key limitation persists: the spatial resolution dilemma. 
Freely accessible missions like Sentinel-2 provide global coverage and high temporal frequency but at a moderate spatial resolution of 10\,m/pixel, while commercial systems like PlanetScope deliver 3\,m/pixel imagery at higher cost and with reduced temporal availability. 
For fine-grained tasks such as burning area mapping or building delineation, this difference limits the quality of the analysis~\cite{acharki2022planetscope,mansaray2021comparing}.

\textbf{Super-Resolution (SR)} attempts to bridge this gap by reconstructing high-resolution (HR) images from low-resolution (LR) observations. 
SR has shown strong benefits in downstream RS tasks~\cite{shermeyer2019effects,jiang2019edge,zhang2014example}, and diffusion-based generative models now represent the state-of-the-art due to their texture recovery and global consistency~\cite{liu20232,yue2023resshift,meng2024conditional,liu2022diffusion}. 

However, satellite super-resolution faces a fundamental challenge: the absence of true LR–HR pairs, since each sensor has a fixed native resolution. As a result, most prior work relies on synthetic intra-sensor degradation (e.g., bicubic downsampling), which fails to capture the complex spectral and radiometric properties of real sensors such as Sentinel-2.

\begin{figure}[t]
    \centering 
    \includegraphics[width=0.80\textwidth]{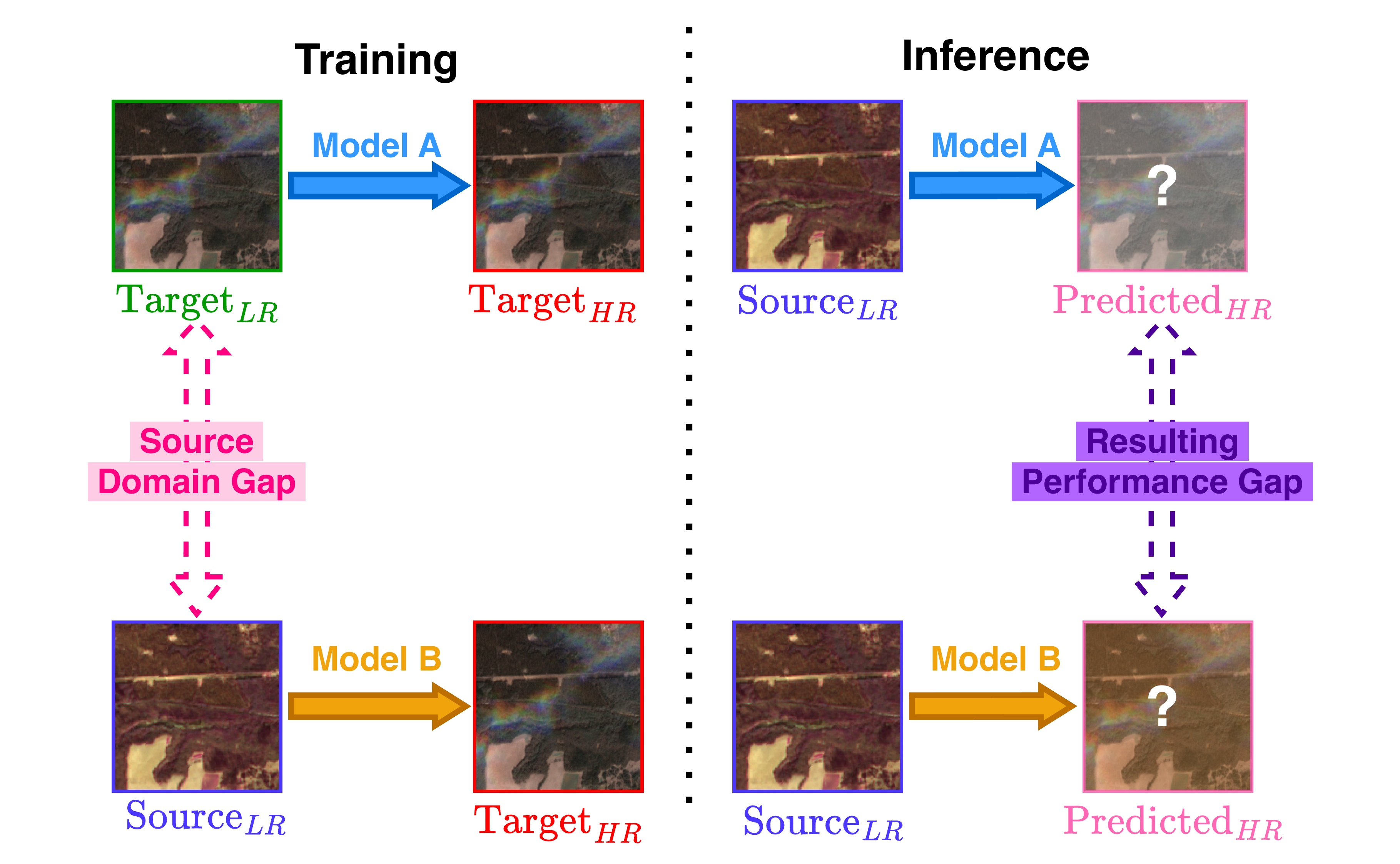}
    \caption{Conceptualizing the cross-sensor SR problem. We compare two distinct models: \textbf{Model A}, trained on the synthetic domain (\textit{Target\_LR}), and 
    \textbf{Model B}, trained on the real-world domain (\textit{Source\_LR}). This paper investigates how the Source Domain Gap (pink) between their training inputs translates into a Resulting Performance Gap (purple) at inference.
    }
    \label{fig:impact}
\end{figure}

In practice, the objective is cross-sensor enhancement, for example, upscaling Sentinel-2 imagery to PlanetScope-like quality. This introduces a substantial domain gap between synthetic training data and real LR observations, caused by differences in spatial resolution, spectral response, and radiometry. As illustrated in Figure~\ref{fig:impact}, models trained on synthetic data often produce unstable results when applied to real satellite imagery.

In this work, we systematically study the \textit{impact of the domain gap} on satellite SR. Specifically, we investigate how diffusion-based SR models trained on synthetic degradations generalise to real cross-sensor scenarios, quantifying performance degradation. Our contributions are threefold: \\
(i) We analyse the \textit{synthetic-to-real} generalisation gap using diverse diffusion-based SR models under consistent conditions. \\
(ii) We introduce a perceptually grounded evaluation framework, including the novel \textit{LPIPS\textsubscript{Sat}} metric based on Sentinel-2 representation learning. \\
(iii) We construct a large-scale paired dataset of Sentinel-2 and PlanetScope imagery, temporally and geometrically aligned to enable rigorous cross-sensor SR evaluation.

We provide the first systematic quantification of how the domain gap influences modern diffusion-based SR, aiming to bridge the gap between synthetic benchmarks and real-world applications.

\section{Related Work}

\textbf{Diffusion Models for Super-Resolution.}
The application of diffusion models~\cite{ho2020denoising,dhariwal2021diffusion} to Super-Resolution (SR) began with SR3~\cite{saharia2022image} and SRDiff~\cite{li2022srdiff}, demonstrating that iterative refinement captures high-fidelity details. Later work introduced upsampling controls~\cite{gao2023implicit}, modifications to the sampling trajectory for improved efficiency~\cite{kopec2025supresdiffgan} and quality~\cite{yue2023resshift}, and more accurate degradation modeling~\cite{liu20232}. Recently, attention shifted to the \textit{blind-SR}~\cite{li2025diffusion} setting to improve performance on real-world data, culminating in powerful prior-guided models like DiffBIR~\cite{lin2024diffbir} and SUPIR \cite{yu2024scaling}.

\textbf{Diffusion in Remote Sensing SR.}
Although diffusion-based SR has achieved remarkable progress in the natural image domains, its application to remote sensing (RS) imagery introduces additional challenges. Satellite data are characterised by large image sizes, multi-spectral channels, and strong geometric variability, making them an ideal, yet demanding, use case for generative models~\cite{qi2025advancing}. Early works such as DMDC~\cite{liu2022diffusion} demonstrated that diffusion frameworks can plausibly reconstruct missing spatial details in satellite imagery. Later studies improved this approach by integrating Transformer-based architectures to capture long-range spatial dependencies~\cite{wu2023hsr}. Parallel research aimed to reduce the high computational burden of diffusion models through lightweight conditioning and efficient sampling strategies~\cite{meng2024conditional}. Recent work targets \textit{blind-SR} setting, where models learn to handle complex sensor-specific degradations representative of real acquisition conditions~\cite{weng2025efficient}. Despite these advances, most existing methods continue to rely on synthetic, intra-sensor degradations, which limits their applicability to operational RS data.

\textbf{Cross-Sensor Domain Gap.}
The central challenge in real-world RS-SR remains the cross-sensor domain gap: models trained on synthetic degradations (e.g., bicubic downsampling) often fail when tested on data from physically distinct sensors. This issue, closely related to the \textit{blind-SR} paradigm, arises from discrepancies in sensor optics, spectral response functions, radiometric calibration, and atmospheric conditions. Recent work has explored several directions to mitigate this problem, from domain-adaptive GANs~\cite{wang2024super} to customised diffusion architectures~\cite{miao2025research}. A comprehensive analysis by Michel \textit{et al.}~\cite{michel2025revisiting} highlighted how real-world geometric and radiometric inconsistencies invalidate standard SR evaluation metrics, highlighting the need for cross-domain benchmarks. However, despite these findings, the impact of this domain gap on modern diffusion-based SR models remains largely unexplored.

In this paper, we address this gap by presenting the first systematic evaluation of diffusion-based super-resolution under cross-sensor conditions. Our benchmark quantifies the synthetic-to-real discrepancy and establishes a foundation for robust cross-sensor generalisation.

\section{Methodology}

\subsection{Dataset}
\label{sec:dataset}

We constructed a dataset of geometrically and temporally aligned image pairs acquired from the \textit{Sentinel-2}~\cite{Drusch2012Sentinel2} and \textit{PlanetScope}~\cite{Planet2025API} satellites. The choice of these two sensors is particularly suitable for this study due to their complementary characteristics. Sentinel-2 provides freely available multispectral imagery at a spatial resolution of 10 m/pixel, which is, however, insufficient for many fine-grained downstream applications. In contrast, PlanetScope offers 3 m/pixel resolution, approaching commercial-grade quality, though it is typically subject to licencing restrictions. Using two satellite systems instead of aerial imagery allows for large-scale coverage while maintaining comparable imaging conditions. The moderate scale factor $(< 4\times)$ makes SR feasible for most modern generative methods, originally developed for much higher scaling factors. Both datasets were acquired through PlanetHub, ensuring sub-pixel geometric alignment consistent with standard L2A processing levels.

We selected 175 PlanetScope regions from USA, each covering approximately $35 \times 25 km$ with cloud coverage below $10\%$. For every region acquired between $15 - 30$ December 2024, a cloud-free Sentinel-2 scene was retrieved. Temporal proximity was prioritised: 154 pairs fall within $\pm15$ days, 10 within $\pm30$ days, and 11 within $\pm45$ days. While temporal offsets affect illumination, we treat these residual misalignments as an integral component of the real-world cross-sensor domain gap rather than noise.

To strictly isolate sensor-specific gaps from scene evolution, we rigorously filtered the test set to include only pairs within a narrow $\pm15$-day window, while the remaining 17 non-overlapping regions were reserved for testing. The geographic distribution is visualised in Figure~\ref{fig:map_patch}.

\begin{figure}[t]
    \centering 
    \includegraphics[width=0.80\textwidth]{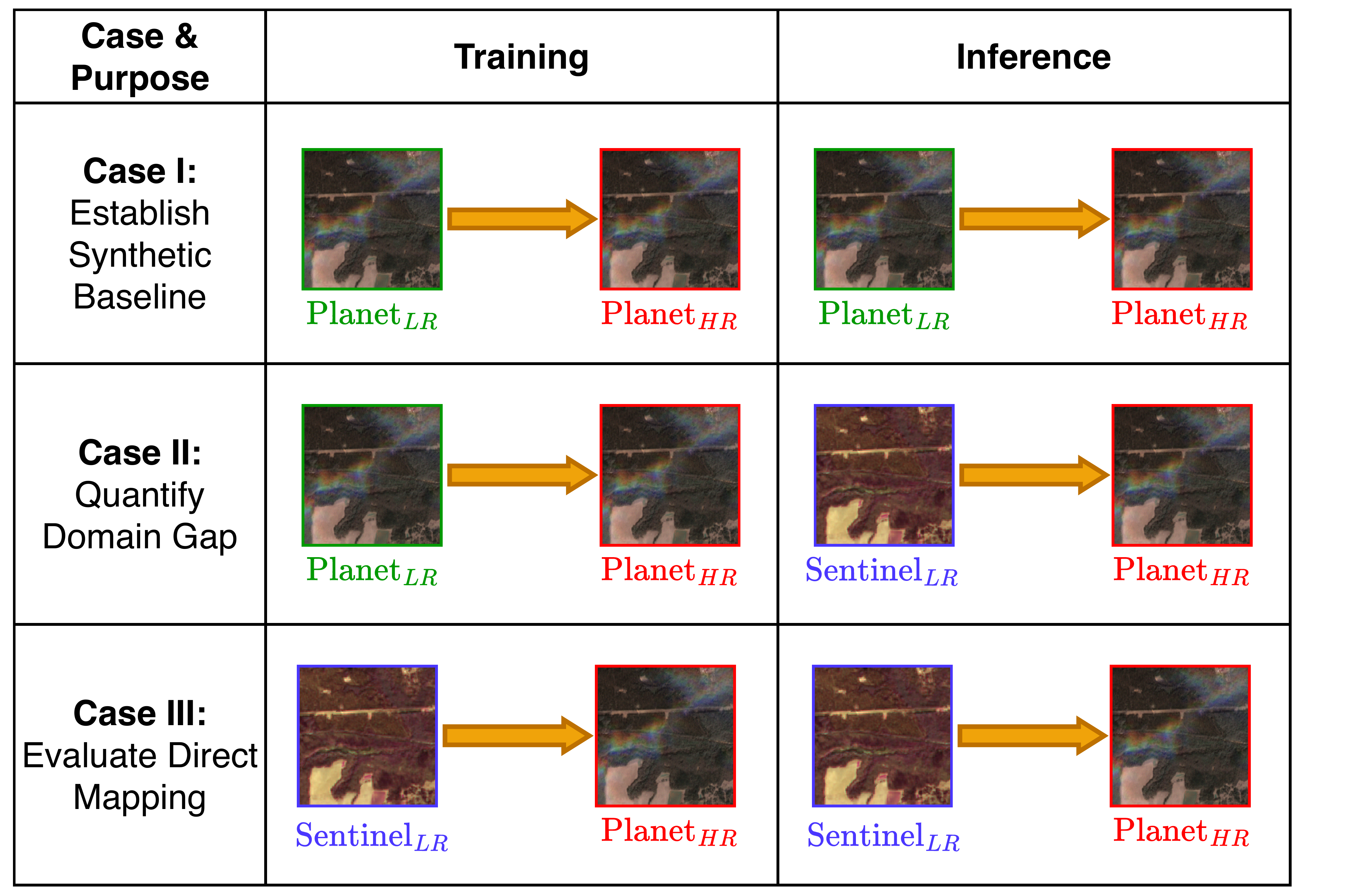}
    \caption{The three experimental configurations. \textbf{Case I} establishes the synthetic baseline. \textbf{Case II} quantifies the synthetic-to-real domain gap. \textbf{Case III} evaluates the alternative direct mapping.}
    \label{fig:cases}
\end{figure}

The resulting dataset serves as the basis for three complementary experimental configurations (Figure~\ref{fig:cases}):

\begin{figure}[t]
    \centering 
    \includegraphics[width=0.95\textwidth]{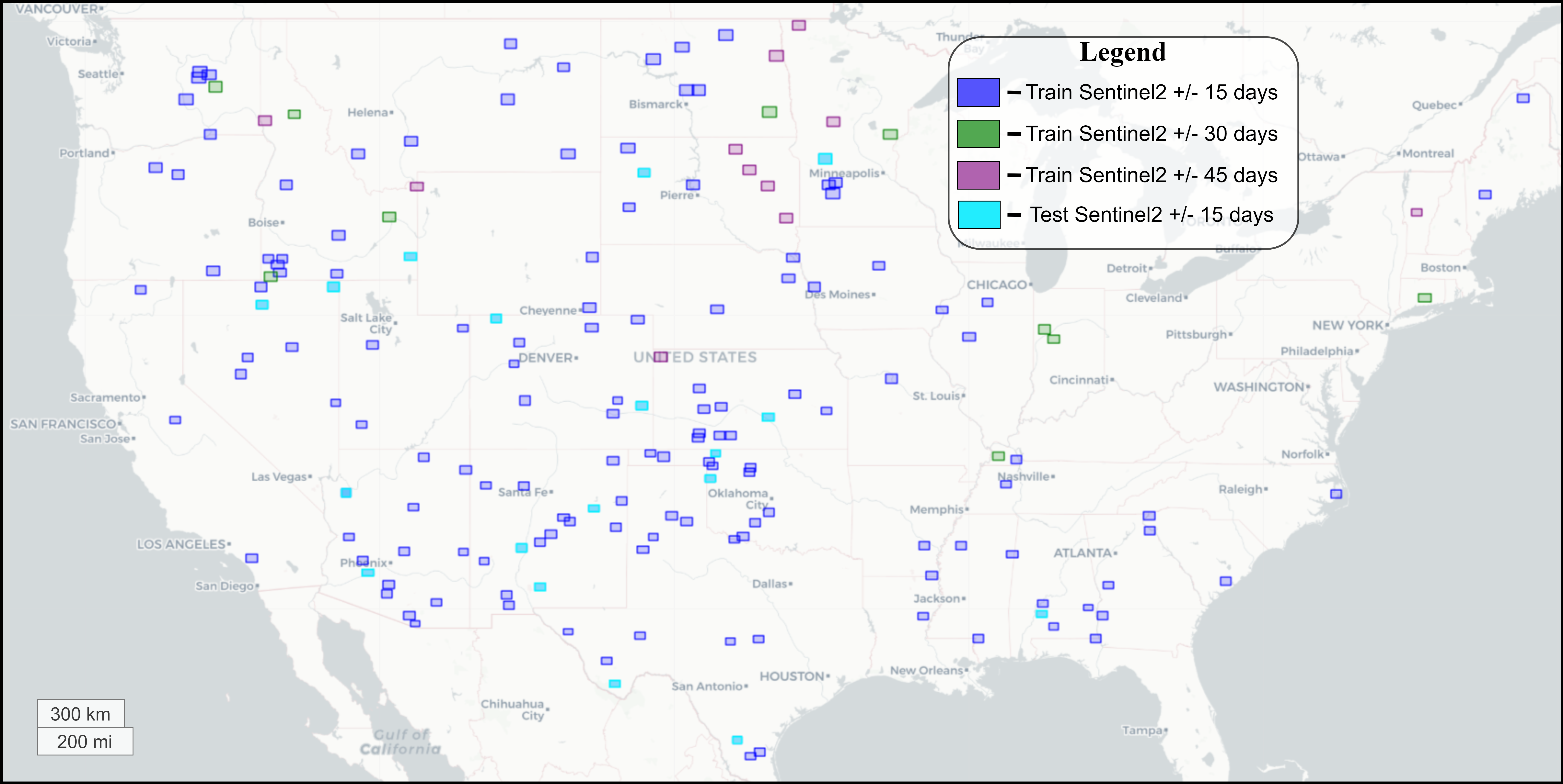}
    \caption{Map of the train/test area divided into collected patches. Sentinel-2 image acquisition time $\pm15, 30 \text{ and } 45$ days from the PlanetScope image time and test data.}
    \label{fig:map_patch}
\end{figure}

\begin{enumerate}
    \item \textbf{Establishing the Synthetic Baseline (Case I):} Training and testing on \textit{Planet\textsubscript{LR}} $\rightarrow$ \textit{Planet\textsubscript{HR}}. This baseline setup represents the common but unrealistic scenario in which models are trained and evaluated on synthetically degraded data.
    \item \textbf{Quantifying the Domain Gap (Case II):} Training on \textit{Planet\textsubscript{LR}} $\rightarrow$ \textit{Planet\textsubscript{HR}} and testing on \textit{Sentinel\textsubscript{LR}} $\rightarrow$ \textit{Planet\textsubscript{HR}}. This configuration measures the real-world impact of the domain gap, highlighting the limitations of training solely on synthetic data.
    \item \textbf{Evaluating Direct Mapping (Case III):} Training and testing on \textit{Sentinel\textsubscript{LR}} $\rightarrow$ \textit{Planet\textsubscript{HR}}. This setup enables evaluating the potential and limitations of directly learning cross-sensor mappings.
\end{enumerate}

\subsection{Benchmarked Diffusion Methods}
\label{sec:methods}

All evaluated models follow the denoising diffusion paradigm~\cite{ho2020denoising}, which defines a \emph{forward} noising process and a learnt \emph{reverse} process parameterized by a neural network (typically a U-Net). Let $x_0$ denote the clean target image, $y_0$ the conditioning low resolution input, $t$ the timestep, and $\epsilon \sim \mathcal{N}(0,\mathbf{I})$ Gaussian noise.
Diffusion models are well-suited for satellite SR because their iterative refinement allows them to recover high-frequency details while maintaining global consistency. Unlike deterministic regression, this iterative refinement enables the generative synthesis of high-frequency details, prioritising perceptual realism over the pixel-average fidelity typical of feed-forward baselines.

\textbf{Diffusion (DDPM)}~\cite{ho2020denoising}
Serves as our baseline. The method trains its network $\epsilon_\theta$ with a simple mean-squared error objective on the noise:

\begin{equation}
    \mathcal{L}_{\text{DDPM}} = \mathbb{E}_{t, x_0, \epsilon \sim \mathcal{N}(0, \mathbf{I})} \left[ \| \epsilon - \epsilon_\theta(x_t, t, y_0) \|^2 \right],
\end{equation}
where,$x_t = \sqrt{\bar{\alpha}_t}x_0 + \sqrt{1-\bar{\alpha}_t}\epsilon$ and $\bar{\alpha}_t$ is predefined variance schedule parameters. DDPM directly models the reverse noising dynamics by estimating the added noise. As a generative model, its ability for domain adaptation is implicit; it must learn complex one-to-many mapping directly from a pixel-level objective.

\textbf{Flow Matching (FM)}~\cite{lipman2022flow}
Represents a deterministic generative paradigm. Instead of learning a score function, Flow Matching trains a deterministic vector field $v_\theta$ that transports samples from a simple prior distribution to the data distribution. The training objective is defined as:

\begin{equation}
    \mathcal{L}_{\text{FM}} = \mathbb{E}_{t, p_t(x_t)} \left[ \| v_\theta(x_t, t, y_0) - u_t(x_t) \|^2 \right],
\end{equation}
where, $x_t = (1 - t)x_0 + t\epsilon$, $p_t(x_t)$ is a target probability density path and $u_t(x_t)$ is the corresponding vector field of this path. By learning a deterministic flow, FM can be more stable and efficient for domain adaptation.

\textbf{Image-to-Image Schrödinger Bridge (I$^2$SB)}~\cite{liu20232}
Frames image-to-image translation as learning a stochastic bridge between conditional distributions, but uses an SGM-style $\epsilon$ parameterisation  for stability~\cite{dhariwal2021diffusion}. The network $\epsilon_\theta$ minimises:

\begin{equation}
    \mathcal{L}_{I^2SB} = \mathbb{E}_{t, \mathbf{x}_0, \epsilon \sim \mathcal{N}(0, \mathbf{I})} \left[ \left\|\epsilon_\theta(\mathbf{x}_t, t, y_0) - \epsilon \right\|^2 \right],
\end{equation}
where, $x_t = \frac{\bar{\sigma}_t^2}{\bar{\sigma}_t^2 + \sigma_t^2} x_0 + \frac{\sigma_t^2}{\bar{\sigma}_t^2 + \sigma_t^2} y_0 + \sqrt{\frac{\sigma_t^2 \bar{\sigma}_t^2}{\bar{\sigma}_t^2 + \sigma_t^2}} \epsilon$, $\sigma_t$ and $\bar{\sigma}_t$ are accumulated noise variance relative to target $x_0$ and condition $y$, respectively. Using this stable, SGM-style objective, I$^2$SB can focus on the generative process (the bridge) rather than on a difficult-to-learn score function, which is ideal for complex cross-domain adaptation.

\textbf{ResShift}~\cite{yue2023resshift}
Re-parameterizes the Markov chain to shift residuals between an HR image and its downsample version (LR). Model trains a network $f_\theta$ to predict the clean HR image directly instead of the noise:

\begin{equation}
    \mathcal{L}_{\text{ResShift}} = \mathbb{E}_{t, x_0} \left[w_t \| f_\theta(x_t, y_0, t) - x_0 \|^2 \right]
\end{equation}
where, $x_t = (1 - \eta_t) x_0 + \eta_t y_0 + \kappa \sqrt{\eta_t} \cdot \epsilon$, $\eta_t$ is shifting sequence, $\kappa$ is a hyper-parameter controlling the noise variance, and $w_t$ are timestep weights. Direct image prediction supplies strong, pixel-level supervision that is helpful for SR and large domain gaps.

\textbf{UniDB}~\cite{zhu2025unidb}
Generalises the concept of diffusion bridges by defining the generative process as a \textit{Stochastic Optimal Control (SOC)}. Instead of just matching a score, the goal is to find an optimal controller $u_{t,\gamma}^*$ that directs the process from the source to the target:

\begin{equation}
    \mathcal{L}_{\text{UniDB}} = \mathbb{E}_{t,\mathbf{x}_0} \left[ \frac{1}{2\sigma_{t-1,\theta}^2} \|\mu_{t-1,\theta} - \mu_{t-1,\gamma}\|_1 \right]
\end{equation}
where, $\mu_{t-1,\theta}$ is the predicted mean of the reverse step $p_\theta(\mathbf{x}_{t-1} | \mathbf{x}_t, \mathbf{x}_T)$ and $\mu_{t-1,\gamma}$ is the target mean derived from the forward process. $\mathbf{x}_t$ is sampled from the bridge posterior $p(\mathbf{x}_t | \mathbf{x}_0, \mathbf{x}_T)$. This allows UniDB to robustly bridge two distinct domains, providing a consistent generative trajectory even under significant cross-sensor or cross-domain shifts.

\medskip
Once trained, each model can sample iteratively by transporting the initial sample $x_T$ to the distribution $x_0$ using trained functions with parameters $\theta$. In experiments configurations \textit{C2} and \textit{C3} we use $x_0$ as \textit{Planet\textsubscript{HR}}, but $y_0$ is different in testing. For all experiments, we use the official, publicly available implementations provided by the authors to ensure a fair and reproducible comparison.

\subsection{Evaluation Metrics}

We evaluated models using four metrics covering both reconstruction fidelity and perceptual quality: PSNR, SSIM, LPIPS, and our proposed satellite-adapted variant \textit{LPIPS\textsubscript{Sat}}.

\textbf{PSNR} measures pixel-wise reconstruction accuracy but is highly sensitive to misalignment and correlates poorly with perceptual quality, particularly in high-frequency remote sensing textures.

\textbf{SSIM} captures luminance, contrast, and structural similarity, offering a more perceptual measure than PSNR, although it still struggles with the complex spatial patterns characteristic of satellite imagery.

\textbf{LPIPS} compares deep feature representations rather than raw pixels, providing a stronger proxy for perceptual similarity. Lower LPIPS values indicate more realistic texture and structural preservation.

\textbf{LPIPS\textsubscript{Sat}} is our domain-adapted metric designed specifically for satellite data. Standard LPIPS relies on ImageNet-pretrained networks, which do not reflect the spectral, radiometric, or textural characteristics of remote sensing imagery. To address this, we employ a ResNet-50~\cite{he2016deep} backbone from \texttt{TorchGeo}, pretrained in a self-supervised manner on Sentinel-2 RGB data~\cite{stewart2023ssl4eo}. To accommodate the 6-channel multispectral data, we compute the metric strictly on the corresponding RGB spectral bands (B4, B3, B2), ensuring alignment with the backbone's pretraining domain. The network remains frozen and for each reference/reconstruction pair (normalised to $[0,1]$), feature maps from \texttt{conv1} and \texttt{layers 1–4} are extracted, channel-wise $L_2$ normalised and compared. Layer-wise feature differences are averaged spatially and then across layers to yield a single perceptual distance.

Using a representation learned directly from Sentinel-2, LPIPS\textsubscript{Sat} provides a perceptual assessment that more faithfully reflects structures and textures present in Earth observation imagery, offering a more meaningful evaluation than generic perceptual metrics.

\subsection{Downstream Task Evaluation}
\label{sec:downstram_task_evaluation}

To assess the practical impact of the domain gap in SR models, we evaluated them on a downstream task: burned-area change delineation. We used the CaBuAr dataset~\cite{cambrin2023cabuar} from \texttt{torchgeo.datasets}, which contains 424 pre-/post-fire Sentinel-2 image pairs ($512\times512$ px) from California wildfires (2015–2022). The data span diverse vegetation types and fire severities, providing a realistic test of cross-sensor robustness.

The downstream model was a U-Net (31M parameters) following the configuration from the CaBuAr paper. Burned-area mapping is moderately complex and highly sensitive to texture and spectral cues—features that SR methods may enhance or distort—while its binary segmentation nature enables clear, quantitative comparison.

Each CaBuAr image was optionally enhanced using the SR models from Section~\ref{sec:methods}; SR was applied independently to pre- and post-fire inputs to isolate its effect.

We evaluated three configurations:
\begin{enumerate}
\item \textbf{Baseline:} Training and testing on original \mbox{Sentinel-2} images.
\item \textbf{SR-tested:} Training on originals, testing on SR-enhanced inputs (measures the domain mismatch introduced by SR).
\item \textbf{SR-trained:} Training and testing on SR-enhanced images (checks whether retraining mitigates the gap).
\end{enumerate}

All setups were trained for 60 epochs using identical hyperparameters and CaBuAr splits. Performance was measured with \textit{IoU} and \textit{F1-score}, standard metrics for binary segmentation.

This evaluation provides a practical view of how SR affects downstream environmental monitoring and whether it preserves task-relevant information.

\section{Experiments}

\subsection{Training Details}
\label{sec:training_details}

All experiments were conducted using the dataset described in Section~\ref{sec:dataset}. 
Each of the $5$ evaluated diffusion-based SR methods were trained under identical conditions to ensure fair comparison. 
All models employed the same U-Net~\cite{ronneberger2015u} backbone with 30M parameters, implemented using the \texttt{diffusers} library~\cite{von2022diffusers}. 
This unified architecture was chosen to isolate the effect of the diffusion formulation itself rather than network capacity or design.

Training was performed for $300,000$ steps, which in preliminary experiments was found sufficient to ensure convergence for all methods given the dataset scale and model size. 
Optimisation was carried out using the AdamW optimizer~\cite{kingma2014adam} with a learning rate of $1\times10^{-4}$, $\eta_{min}=1\times10^{-7}$, and a cosine annealing scheduler. The batch size was set to 4, and training was performed on NVIDIA A100 GPUs in mixed-precision mode to balance efficiency and numerical stability. No data augmentation was applied to preserve the spectral and spatial consistency of the paired satellite imagery.

The validation was conducted every 10k iterations, and the final checkpoint was selected based on the best \textit{LPIPS\textsubscript{Sat}} score on the validation set.  For inference, all models used a standardised budget of $T=1000$ steps with default schedulers and no classifier-free guidance.

\subsection{Data Preparation}

All datasets were processed to ensure geometric, temporal, and spatial consistency across sensors. First, both Sentinel-2 and PlanetScope scenes were reprojected to a common coordinate reference system (EPSG:4326). Sub-pixel geometric alignment was achieved through affine coregistration.

Each PlanetScope image was divided into $512\times512$ tiles using a $25\%$ overlap. For every high-resolution tile (\textit{Planet\textsubscript{HR}}), a corresponding Sentinel-2 patch was extracted and downsampled by a factor of $\frac{10}{3}$ to match spatial resolution. Tile pairs containing more than $10\%$ missing pixels (zero-valued across all channels) were removed.

To support diffusion-based SR models, all Sentinel-2 patches were upsampled to $512\times512$ using bicubic interpolation, forming the \textit{Sentinel\textsubscript{LR}} set. For synthetic degradation experiments, low-resolution PlanetScope tiles (\textit{Planet\textsubscript{LR}}) were created by downsampling \textit{Planet\textsubscript{HR}} to Sentinel-2 scale using the \texttt{rasterio} library—ensuring geospatial consistency—and subsequently upsampling back to $512\times512$ using bicubic interpolation.

The final dataset comprises $61,239$ training pairs and $7,303$ test pairs, derived exclusively from the $\pm15$-day subset to reduce temporal variability in the evaluation stage.

A key challenge in cross-sensor tasks is spectral alignment, as sensors capture data in distinct bands corresponding to specific wavelengths. Sentinel-2 (12 bands) and PlanetScope (8 bands) do not overlap perfectly. To ensure a physically valid SR mapping, we selected six bands with matching spectral properties, as detailed in Table~\ref{tab:band_overlap}. Using non-overlapping bands would force the model to "hallucinate" spectral information (e.g., predict Red from Blue), which is not a valid SR task.

\begin{table}[t]
\centering
\caption{Spectral band alignment for cross-sensor experiments. We selected six bands with overlapping spectral wavelengths ($\lambda$) to ensure a physically consistent mapping.}
\begin{tabular}{|c|c|c|c|}
\hline
\textbf{Common Name} & \textbf{Wavelength ($\lambda$)} & \begin{tabular}{@{}c@{}}\textbf{PlanetScope} \\ \textbf{Band ID}\end{tabular} & \begin{tabular}{@{}c@{}}\textbf{Sentinel-2} \\ \textbf{Band ID}\end{tabular} \\ \hline
Coastal Blue & $\sim$443 nm& 1 & 1\\ \hline
Blue & $\sim$490 nm& 2 & 2 \\ \hline
Green& $\sim$560 nm & 4 & 3\\ \hline
Red& $\sim$665 nm & 6 & 4\\ \hline
Red Edge & $\sim$705 nm & 7 & 5 \\ \hline
Near-Infrared (NIR) & $\sim$865 nm & 8 & 8A \\ \hline
\end{tabular}
\label{tab:band_overlap}
\end{table}

\subsection{Experimental Results}

\begin{table*}[t]
\caption{
    Quantitative metric comparison. \textit{C1}: synthetic (\textit{Planet}$\rightarrow$\textit{Planet}).\textit{C2}: cross-sensor generalization (tested on \textit{Sentinel}). \textit{C3}: cross-sensor adaptation (trained on \textit{Sentinel}). \textit{Default}: metrics between low-resolution inputs and high-resolution ground truth. LPIPS\textsubscript{Sat} results are scaled by $ (\times 10^{-3}) $. The best and second-best results are highlighted in \textbf{bold} and \underline{underline}, respectively. 
}
\begin{center}
\resizebox{\textwidth}{!}{
\begin{tabular}{|c!{\vrule width 1.5pt}ccc!{\vrule width 1.5pt}ccc!{\vrule width 1.5pt}ccc!{\vrule width 1.5pt}ccc!{\vrule width 1.5pt}}
\hline
\textbf{Metric}             & \multicolumn{3}{c!{\vrule width 1.5pt}}{\textit{\textbf{PSNR \(\uparrow\)}}}                 & \multicolumn{3}{c!{\vrule width 1.5pt}}{\textit{\textbf{SSIM \(\uparrow\)}}}                & \multicolumn{3}{c!{\vrule width 1.5pt}}{\textit{\textbf{LPIPS \(\downarrow\)}}}              & \multicolumn{3}{c!{\vrule width 1.5pt}}{\textit{\textbf{LPIPS\textsubscript{Sat} \(\downarrow\)}}}                \\ \hline
\textbf{Case}               & \multicolumn{1}{c|}{\textit{\textbf{C1}}}                & \multicolumn{1}{c|}{\textit{\textbf{C2}}}               & \textit{\textbf{C3}}                & \multicolumn{1}{c|}{\textit{\textbf{C1}}}               & \multicolumn{1}{c|}{\textit{\textbf{C2}}}               & \textit{\textbf{C3}}               & \multicolumn{1}{c|}{\textit{\textbf{C1}}}               & \multicolumn{1}{c|}{\textit{\textbf{C2}}}               & \textit{\textbf{C3}}               & \multicolumn{1}{c|}{\textit{\textbf{C1}}}               & \multicolumn{1}{c|}{\textit{\textbf{C2}}}               & \textit{\textbf{C3}}               \\ \hline
\textit{\textbf{Diffusion}} & \multicolumn{1}{c|}{\underline{45.26}} & \multicolumn{1}{c|}{16.71}                             & 20.83                              & \multicolumn{1}{c|}{\underline{0.98}} & \multicolumn{1}{c|}{0.61}                              & 0.82                              & \multicolumn{1}{c|}{0.04}                              & \multicolumn{1}{c|}{0.39}                              & 0.37                              & \multicolumn{1}{c|}{0.37}                              & \multicolumn{1}{c|}{2.21}                              & 1.96                              \\ \hline
\textit{\textbf{I$^2$SB}}   & \multicolumn{1}{c|}{45.20}                              & \multicolumn{1}{c|}{\underline{16.72}} & 22.39                              & \multicolumn{1}{c|}{\underline{0.98}} & \multicolumn{1}{c|}{\underline{0.62}} & \textbf{0.85}                     & \multicolumn{1}{c|}{\underline{0.03}}                              & \multicolumn{1}{c|}{\underline{0.37}} & \underline{0.27} & \multicolumn{1}{c|}{\textbf{0.34}}                     & \multicolumn{1}{c|}{\underline{2.15}} & \textbf{1.53}                     \\ \hline
\textit{\textbf{ResShift}}  & \multicolumn{1}{c|}{\textbf{45.58}}                     & \multicolumn{1}{c|}{\textbf{16.73}}                    & 16.88                              & \multicolumn{1}{c|}{\textbf{0.99}}                     & \multicolumn{1}{c|}{\textbf{0.63}}                     & 0.64                              & \multicolumn{1}{c|}{0.04}                              & \multicolumn{1}{c|}{\textbf{0.36}}                     & 0.30                             & \multicolumn{1}{c|}{0.36}                              & \multicolumn{1}{c|}{\textbf{2.14}}                     & 1.90                              \\ \hline
\textit{\textbf{FM}}        & \multicolumn{1}{c|}{44.72}                              & \multicolumn{1}{c|}{16.70}                             & \underline{22.56} & \multicolumn{1}{c|}{0.97}                              & \multicolumn{1}{c|}{0.61}                              & 0.83                              & \multicolumn{1}{c|}{\underline{0.03}} & \multicolumn{1}{c|}{0.39}                              & 0.28                              & \multicolumn{1}{c|}{0.37}                              & \multicolumn{1}{c|}{2.18}                              & \underline{1.60} \\ \hline
\textit{\textbf{Unidb}}     & \multicolumn{1}{c|}{44.41}                              & \multicolumn{1}{c|}{16.70}                             & \textbf{23.81}                     & \multicolumn{1}{c|}{0.97}                              & \multicolumn{1}{c|}{0.61}                              & \underline{0.84} & \multicolumn{1}{c|}{\textbf{0.02}}                     & \multicolumn{1}{c|}{0.38}                              & \textbf{0.26}                     & \multicolumn{1}{c|}{\underline{0.35}} & \multicolumn{1}{c|}{2.18}                              & 1.63                              \\ \hline
\textit{\textbf{Default}}   & \multicolumn{1}{c|}{43.55}              & \multicolumn{2}{c!{\vrule width 1.5pt}}{16.88}                    & \multicolumn{1}{c|}{0.97}               & \multicolumn{2}{c!{\vrule width 1.5pt}}{0.64}                           & \multicolumn{1}{c|}{0.10}               & \multicolumn{2}{c!{\vrule width 1.5pt}}{0.30}                           & \multicolumn{1}{c|}{0.45}               & \multicolumn{2}{c!{\vrule width 1.5pt}}{1.89}                           \\ \hline
\end{tabular}
}
\end{center}
\label{tab:metrics}
\end{table*}

\begin{figure}[t]
    \centering
    
    \begin{subfigure}[t]{0.16\textwidth} 
         \centering 
         \includegraphics[width=\linewidth, trim=20pt 90pt 130pt 60pt, clip=true]{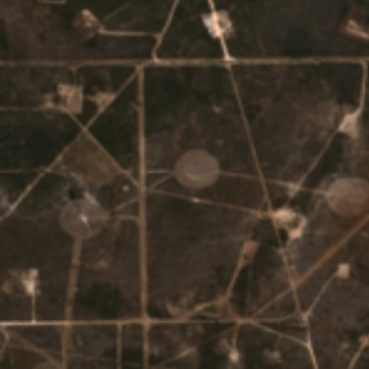}
         \begin{center}\scriptsize\textbf{Sentinel$_{LR}$}\end{center}
    \end{subfigure}
    \hfill
    \begin{subfigure}[t]{0.16\textwidth}
         \centering
         \includegraphics[width=\linewidth, trim=20pt 90pt 130pt 60pt, clip=true]{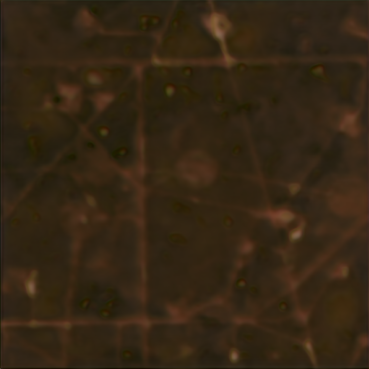}
         \begin{center}\scriptsize\textbf{Diffusion \\ Sentinel}\end{center}
    \end{subfigure}
    \hfill
    \begin{subfigure}[t]{0.16\textwidth}
         \centering
         \includegraphics[width=\linewidth, trim=20pt 90pt 130pt 60pt, clip=true]{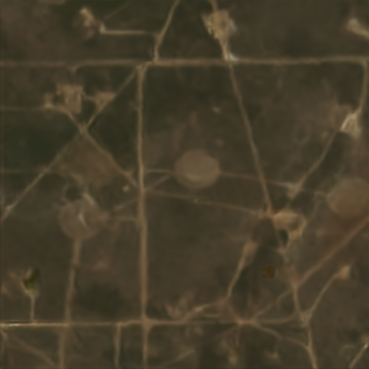}
         \begin{center}\scriptsize\textbf{I$^2$SB \\ Sentinel}\end{center}
    \end{subfigure}
    \hfill
    \begin{subfigure}[t]{0.16\textwidth}
         \centering
         \includegraphics[width=\linewidth, trim=20pt 90pt 130pt 60pt, clip=true]{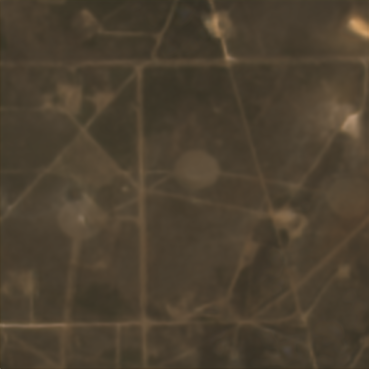}
         \begin{center}\scriptsize\textbf{ResShift \\ Sentinel}\end{center}
    \end{subfigure}
    \hfill
    \begin{subfigure}[t]{0.16\textwidth}
         \centering
         \includegraphics[width=\linewidth, trim=20pt 90pt 130pt 60pt, clip=true]{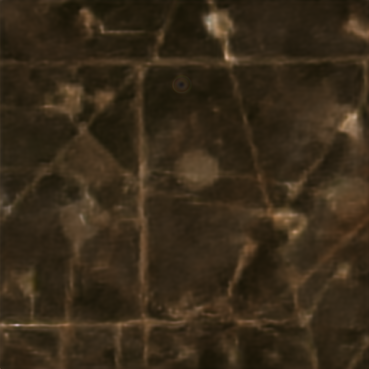}
         \begin{center}\scriptsize\textbf{Unidb \\ Sentinel}\end{center}
    \end{subfigure}
    \hfill
    \begin{subfigure}[t]{0.16\textwidth}
         \centering
         \includegraphics[width=\linewidth, trim=20pt 90pt 130pt 60pt, clip=true]{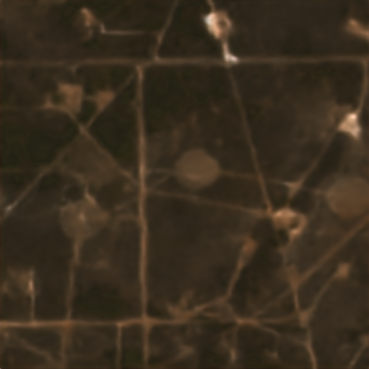}
         \begin{center}\scriptsize\textbf{FM \\ Sentinel}\end{center}
    \end{subfigure}

    \vspace{2mm} 

    \begin{subfigure}[t]{0.16\textwidth}
         \centering
         \includegraphics[width=\linewidth, trim=20pt 90pt 130pt 60pt, clip=true]{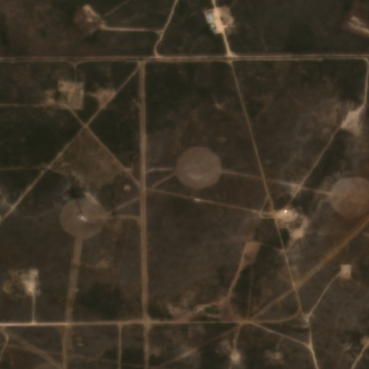}
         \begin{center}\scriptsize\textbf{Planet$_{HR}$}\end{center}
    \end{subfigure}
    \hfill
    \begin{subfigure}[t]{0.16\textwidth}
         \centering
         \includegraphics[width=\linewidth, trim=20pt 90pt 130pt 60pt, clip=true]{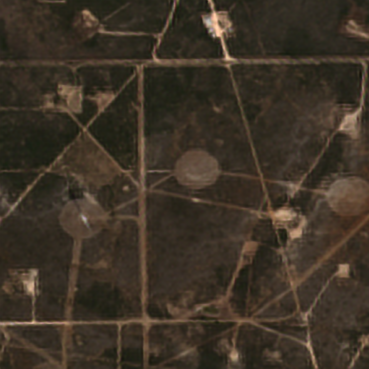}
         \begin{center}\scriptsize\textbf{Diffusion \\ Planet}\end{center}
    \end{subfigure}
    \hfill
    \begin{subfigure}[t]{0.16\textwidth}
         \centering
         \includegraphics[width=\linewidth, trim=20pt 90pt 130pt 60pt, clip=true]{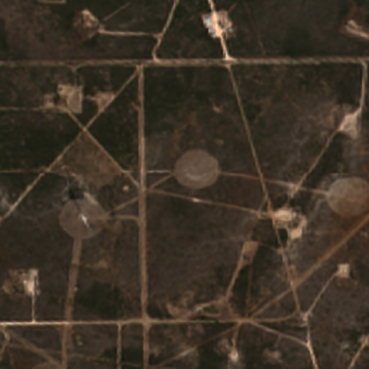}
         \begin{center}\scriptsize\textbf{I$^2$SB \\ Planet}\end{center}
    \end{subfigure}
    \hfill
    \begin{subfigure}[t]{0.16\textwidth}
         \centering
         \includegraphics[width=\linewidth, trim=20pt 90pt 130pt 60pt, clip=true]{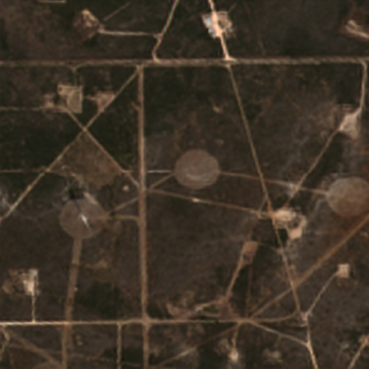}
         \begin{center}\scriptsize\textbf{ResShift \\ Planet}\end{center}
    \end{subfigure}
    \hfill
    \begin{subfigure}[t]{0.16\textwidth}
         \centering
         \includegraphics[width=\linewidth, trim=20pt 90pt 130pt 60pt, clip=true]{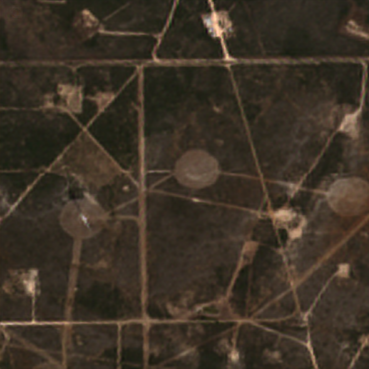}
         \begin{center}\scriptsize\textbf{Unidb \\ Planet}\end{center}
    \end{subfigure}
    \hfill
    \begin{subfigure}[t]{0.16\textwidth}
         \centering
         \includegraphics[width=\linewidth, trim=20pt 90pt 130pt 60pt, clip=true]{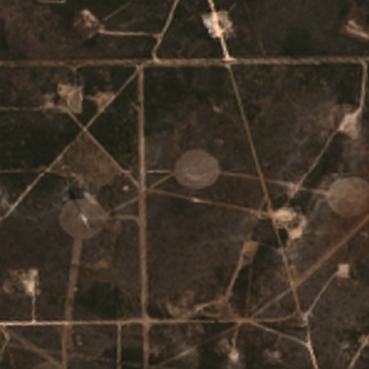}
         \begin{center}\scriptsize\textbf{FM \\ Planet}\end{center}
    \end{subfigure}
    \caption{Visual comparison for the cross-sensor task (zoomed in image). (Top row) \textit{Sentinel\textsubscript{LR}} input (left) and \textit{C3} (real-trained) model outputs. (Bottom row) \textit{Planet\textsubscript{HR}} ground truth (left) and \textit{C2} (synthetic-trained) model outputs.}
    \label{fig:visualization}
\end{figure}

\begin{center}
\begin{table*}[t]
\centering
\caption{Metric variability across Sentinel–Planet test pairs, quantifying the gap between low-resolution inputs and high-resolution ground truth.}
\begin{tabular}{|c|c|c|c|c|}
\hline
\textbf{Metric}       & \textit{\textbf{PSNR \(\uparrow\)}} & \textit{\textbf{SSIM \(\uparrow\)}} & \textit{\textbf{LPIPS \(\downarrow\)}} & \textit{\textbf{LPIPS\textsubscript{Sat} \(\downarrow\)}} \\ \hline
\textit{\textbf{Min}} & 2.306                  & -0.893                 & 0.021                   & 0.567                        \\ \hline
\textit{\textbf{Max}} & 44.256                 & 0.988                  & 1.061                   & 5.388                        \\ \hline
\textit{\textbf{Mean}} & 25.135                 & 0.642                 & 0.295                   & 1.888                        \\ \hline
\textit{\textbf{Std}} & 11.729                & 0.511                 & 0.274                   & 1.114                        \\ \hline
\end{tabular}
\label{tab:domain_gap_variability}
\end{table*}
\end{center}

The quantitative results are summarised in Table~\ref{tab:metrics}, and the performance of the models can be seen in Figure \ref{fig:visualization}. 
The \textit{Default} row provides a reference for both the synthetic (\textit{C1}) and real (\textit{C2, C3}) gaps.

In the synthetic case (\textit{Planet\textsubscript{LR}} $\rightarrow$ \textit{Planet\textsubscript{HR}}, \textit{C1}), the gap is minimal—PSNR exceeds 43\,dB and SSIM reaches 0.97—indicating that artificial degradation produces a simplistic learning setup. In this regime, the reconstructions are nearly indistinguishable from ground truth, both quantitatively and perceptually.

The situation changes dramatically in the real cross-sensor setting. Even the \textit{Default} comparison (\textit{Sentinel\textsubscript{LR}} $\rightarrow$ \textit{Planet\textsubscript{HR}}), which measures \textit{the domain gap}, shows a substantial drop (PSNR $\approx$ 16.9\,dB, LPIPS = 0.296). More importantly, this gap is highly non-uniform. As shown in Table~\ref{tab:domain_gap_variability}, the variability across all metrics is extremely high. This confirms that real-world discrepancies in alignment, illumination, and sensor characteristics introduce significant, unpredictable variability, posing a major challenge for any consistent learning model.

When diffusion models trained on synthetic data are applied to real Sentinel imagery (\textit{C2}), they fail catastrophically. Performance deteriorates even further, falling below the \textit{Default} baseline in both PSNR and SSIM. Perceptual scores (LPIPS/LPIPS\textsubscript{Sat}) also confirm that the models are \textit{actively degrading} the input imagery, likely by imposing false synthetic-data priors (e.g., bicubic-like sharpness) onto physically different Sentinel data. This clearly demonstrates that synthetic training does not provide a meaningful preparation for real cross-sensor challenges.

Training and testing on real data (\textit{C3}) allow partial recovery. Perceptual metrics improve significantly over the \textit{C2} failure, indicating limited domain adaptation. However, high variance in sensor-data (Table~\ref{tab:domain_gap_variability}) hinders consistent optimization across architectures. As a result, performance remains far below the synthetic setting (\textit{C1}), showing that even in-domain data do not produce a stable, physically consistent mapping.

To better capture these perceptual failures, our proposed \textit{LPIPS\textsubscript{Sat}} metric follows trends similar to LPIPS but offers higher sensitivity to the spectral and structural discrepancies unique to remote sensing. For example, in \textit{C3}, \textit{I$^2$SB} achieves the lowest LPIPS\textsubscript{Sat} distance (1.532), indicating that it best preserves spatial and spectral realism under these challenging real-world conditions.

Overall, these findings confirm two distinct challenges. First, a severe domain gap (\textit{C1 vs.\ C2}) that causes synthetically trained models to fail completely. Second, even when this is mitigated, a persistent performance gap (\textit{C1 vs.\ C3}) remains, suggesting that current SR architectures cannot fully adapt to the physical and radiometric diversity of real-world sensors.

\subsection{Downstream Task Results}

\begin{center}
\begin{table*}[t]
\caption{
    Downstream task performance (IoU/F1-score $\uparrow$).
    \textit{SR-tested}: U-Net trained on baseline, tested on SR images.
    \textit{SR-trained}: U-Net trained and tested on SR images.
    \textit{Default}: Baseline performance (no SR).
    \textit{C2}/\textit{C3} indicate which SR model outputs were used.
}
\begin{center}
\begin{tabular}{|c|cccc|cccc|}
\hline
\textbf{Metric}             & \multicolumn{4}{c|}{\textit{\textbf{IoU \(\uparrow\)}}}                                                                                                                                                          & \multicolumn{4}{c|}{\textit{\textbf{F1 \(\uparrow\)}}}                                                                                                                                                           \\ \hline
\textbf{Variant}            & \multicolumn{2}{c|}{\textit{\textbf{SR-tested}}}                                                                         & \multicolumn{2}{c|}{\textit{\textbf{SR-trained}}}                                                    & \multicolumn{2}{c|}{\textit{\textbf{SR-tested}}}                                                                         & \multicolumn{2}{c|}{\textit{\textbf{SR-trained}}}                                                    \\ \hline
\textbf{Case}               & \multicolumn{1}{c|}{\textit{\textbf{C2}}}               & \multicolumn{1}{c|}{\textit{\textbf{C3}}}               & \multicolumn{1}{c|}{\textit{\textbf{C2}}}               & \textit{\textbf{C3}}               & \multicolumn{1}{c|}{\textit{\textbf{C2}}}               & \multicolumn{1}{c|}{\textit{\textbf{C3}}}               & \multicolumn{1}{c|}{\textit{\textbf{C2}}}               & \textit{\textbf{C3}}               \\ \hline
\textit{\textbf{Diffusion}} & \multicolumn{1}{c|}{0.468}                              & \multicolumn{1}{c|}{0.097}                              & \multicolumn{1}{c|}{0.412}                              & \textbf{0.419}                     & \multicolumn{1}{c|}{0.538}                              & \multicolumn{1}{c|}{0.139}                              & \multicolumn{1}{c|}{0.493}                              & \textbf{0.517}                     \\ \hline
\textit{\textbf{I$^2$SB}}   & \multicolumn{1}{c|}{0.463}                              & \multicolumn{1}{c|}{0.035}                              & \multicolumn{1}{c|}{0.419}                              & 0.280                              & \multicolumn{1}{c|}{0.539}                              & \multicolumn{1}{c|}{0.057}                              & \multicolumn{1}{c|}{0.499}                              & 0.354                              \\ \hline
\textit{\textbf{ResShift}}  & \multicolumn{1}{c|}{\textbf{0.473}}                     & \multicolumn{1}{c|}{0.267}                              & \multicolumn{1}{c|}{\underline{0.446}} & \underline{0.354} & \multicolumn{1}{c|}{\underline{0.541}} & \multicolumn{1}{c|}{\underline{0.330}} & \multicolumn{1}{c|}{\underline{0.524}} & 0.403                              \\ \hline
\textit{\textbf{FM}}        & \multicolumn{1}{c|}{\underline{0.470}} & \multicolumn{1}{c|}{\textbf{0.327}}                     & \multicolumn{1}{c|}{\textbf{0.448}}                     & 0.353                              & \multicolumn{1}{c|}{\textbf{0.550}}                     & \multicolumn{1}{c|}{\textbf{0.403}}                     & \multicolumn{1}{c|}{\underline{0.524}} & \underline{0.410} \\ \hline
\textit{\textbf{Unidb}}     & \multicolumn{1}{c|}{0.453}                              & \multicolumn{1}{c|}{\underline{0.270}} & \multicolumn{1}{c|}{0.440}                              & 0.279                              & \multicolumn{1}{c|}{0.527}                              & \multicolumn{1}{c|}{0.319}                              & \multicolumn{1}{c|}{\textbf{0.526}}                     & 0.340                              \\ \hline
\textit{\textbf{Default}}   & \multicolumn{4}{c|}{0.437}                                                                                                                                                                                       & \multicolumn{4}{c|}{0.516}                                                                                                                                                                                       \\ \hline
\end{tabular}
\end{center}
\label{tab:dt_task}
\end{table*}
\end{center}

To quantify the practical impact of the failures at the metric-level, we evaluated the SR-enhanced images on a burnt-area segmentation task. The results are summarised in Table~\ref{tab:dt_task}. We used the \textit{Baseline} U-Net performance as a reference.

First, we analysed the \textit{SR-tested} configuration, which measures domain mismatch. When using models from the synthetic case \textit{C2}, the downstream performance improves compared to the \textit{Baseline}. This shows that SR works and is able not only to improve the image visually but also to help the downstream task. However, when using models from the real-world case \textit{C3}, performance decreased. This may be due to the fact that when learning to project onto the Planet domain, the models also had to learn domain adaptation (Sentinel$\rightarrow$Planet), which is a difficult combined task. This unstable, dual-objective process creates unrealistic artifacts that interfere with the model's understanding of the data.

Next, in the \textit{SR-trained} configuration, the downstream model is retrained on the SR-enhanced images. In this case, the performance gap between \textit{C2} and \textit{C3} narrows, and both configurations achieve results close to the \textit{Baseline}, but do not exceed it significantly or consistently. This modest result may be due to the limitations of the CaBuAr labels; it is plausible that the SR models predict high-resolution details more accurately than the original ground-truth masks and are therefore unfairly penalised for outperforming the labels.

Practically, the trade-off is clear. Synthetically trained models (\textit{C2}) focus on pure SR, offering consistent but modest gains. In contrast, models trained on real data (\textit{C3}) attempt complex domain adaptation; while this improves perceptual metrics, it introduces hallucinations that degrade downstream utility. This divergence confirms that higher SR metrics do not guarantee practical effectiveness. It also explains the persistence of synthetic training: despite the performance gap, it offers a safer, artifact-free baseline compared to the unpredictability of real-world adaptation. \\

\section{Conclusion}

In this paper, we conducted a comprehensive analysis of the synthetic-to-real generalisation gap using several diffusion-based SR models. To do this, we constructed a new large-scale aligned dataset of Sentinel-2 and PlanetScope imagery and introduced a perceptually-grounded evaluation framework, including the novel LPIPS\textsubscript{Sat} metric.

Our results confirm two distinct challenges. First, a domain gap (\textit{C1} vs. \textit{C2}) that causes synthetically-trained models to fail traditional metric evaluations. Second, even when this is mitigated by training on real-world data, a persistent performance gap (\textit{C1} vs. \textit{C3}) remains, suggesting that current architectures cannot fully adapt to the physical and radiometric diversity of real-world sensors.

Most importantly, we demonstrated the practical implications of these gaps on a downstream task. We found that the synthetically-trained \textit{C2} models, by learning a pure SR task, provide a consistent and modest benefit. Conversely, the real-world-trained \textit{C3} models, by being forced to learn an unstable, entangled task of SR and domain adaptation, introduce destructive artifacts that render them unusable.

This finding explains a key dilemma in the field: synthetic training is still commonly used because it is stable, but its utility is ultimately limited. Our work shows that to bridge the performance gap, future research must focus not just on model architectures, but on new techniques to separate domain adaptation from the super-resolution task.

\subsection*{Acknowledgment}

\footnotesize{
We gratefully acknowledge Polish high-performance computing infrastructure PLGrid (HPC Center: ACK Cyfronet AGH) for providing computer facilities and support within computational grant no. PLG/2025/018843

\noindent
The work conducted by Wojciech Kozłowski and Maciej Zięba was supported by the National Centre of Science (Poland) grant no. 2021/43/B/ST6/02853. 
}

\bibliographystyle{splncs04}
\bibliography{bibliography}

@article{stewart2023ssl4eo,
  title={Ssl4eo-l: Datasets and foundation models for landsat imagery},
  author={Stewart, Adam and Lehmann, Nils and Corley, Isaac and Wang, Yi and Chang, Yi-Chia and Ait Ali Braham, Nassim Ait and Sehgal, Shradha and Robinson, Caleb and Banerjee, Arindam},
  journal={Advances in Neural Information Processing Systems},
  volume={36},
  pages={59787--59807},
  year={2023}
}

@inproceedings{he2016deep,
  title={Deep residual learning for image recognition},
  author={He, Kaiming and Zhang, Xiangyu and Ren, Shaoqing and Sun, Jian},
  booktitle={Proceedings of the IEEE conference on computer vision and pattern recognition},
  pages={770--778},
  year={2016}
}

@article{cambrin2023cabuar,
  title={CaBuAr: California burned areas dataset for delineation [Software and Data Sets]},
  author={Cambrin, Daniele Rege and Colomba, Luca and Garza, Paolo},
  journal={IEEE Geoscience and Remote Sensing Magazine},
  volume={11},
  number={3},
  pages={106--113},
  year={2023},
  publisher={IEEE}
}

@article{ma2019deep,
  title={Deep learning in remote sensing applications: A meta-analysis and review},
  author={Ma, Lei and Liu, Yu and Zhang, Xueliang and Ye, Yuanxin and Yin, Gaofei and Johnson, Brian Alan},
  journal={ISPRS journal of photogrammetry and remote sensing},
  volume={152},
  pages={166--177},
  year={2019},
  publisher={Elsevier}
}

@article{shanmugapriya2019applications,
  title={Applications of remote sensing in agriculture-A Review},
  author={Shanmugapriya, Palanisamy and Rathika, Selvaraj and Ramesh, Thanakkan and Janaki, Ponnusamy},
  journal={Int. J. Curr. Microbiol. Appl. Sci},
  volume={8},
  number={01},
  pages={2270--2283},
  year={2019}
}

@article{sun2024application,
  title={Application of remote sensing technology in water quality monitoring: From traditional approaches to artificial intelligence},
  author={Sun, Yuan and Wang, Denghui and Li, Lei and Ning, Rongsheng and Yu, Shuili and Gao, Naiyun},
  journal={Water Research},
  volume={267},
  pages={122546},
  year={2024},
  publisher={Elsevier}
}

@article{acharki2022planetscope,
  title={PlanetScope contributions compared to Sentinel-2, and Landsat-8 for LULC mapping},
  author={Acharki, Siham},
  journal={Remote Sensing Applications: Society and Environment},
  volume={27},
  pages={100774},
  year={2022},
  publisher={Elsevier}
}

@article{mansaray2021comparing,
  title={Comparing PlanetScope to Landsat-8 and Sentinel-2 for sensing water quality in reservoirs in agricultural watersheds},
  author={Mansaray, Abubakarr S and Dzialowski, Andrew R and Martin, Meghan E and Wagner, Kevin L and Gholizadeh, Hamed and Stoodley, Scott H},
  journal={Remote Sensing},
  volume={13},
  number={9},
  pages={1847},
  year={2021},
  publisher={MDPI}
}

@inproceedings{shermeyer2019effects,
  title={The effects of super-resolution on object detection performance in satellite imagery},
  author={Shermeyer, Jacob and Van Etten, Adam},
  booktitle={Proceedings of the IEEE/CVF Conference on Computer Vision and Pattern Recognition Workshops},
  pages={0--0},
  year={2019}
}

@article{liu20232,
  title={{I$^2$SB: Image-to-Image Schr\"{o}dinger Bridge}},
  author={Liu, Guan-Horng and Vahdat, Arash and Huang, De-An and Theodorou, Evangelos A and Nie, Weili and Anandkumar, Anima},
  journal={arXiv preprint arXiv:2302.05872},
  year={2023}
}

@article{jiang2019edge,
  title={Edge-enhanced GAN for remote sensing image superresolution},
  author={Jiang, Kui and Wang, Zhongyuan and Yi, Peng and Wang, Guangcheng and Lu, Tao and Jiang, Junjun},
  journal={IEEE Transactions on Geoscience and Remote Sensing},
  volume={57},
  number={8},
  pages={5799--5812},
  year={2019},
  publisher={IEEE}
}

@article{zhang2014example,
  title={Example-based super-resolution land cover mapping using support vector regression},
  author={Zhang, Yihang and Du, Yun and Ling, Feng and Fang, Shiming and Li, Xiaodong},
  journal={IEEE Journal of Selected Topics in Applied Earth Observations and Remote Sensing},
  volume={7},
  number={4},
  pages={1271--1283},
  year={2014},
  publisher={IEEE}
}

@article{yue2023resshift,
  title={Resshift: Efficient diffusion model for image super-resolution by residual shifting},
  author={Yue, Zongsheng and Wang, Jianyi and Loy, Chen Change},
  journal={Advances in Neural Information Processing Systems},
  volume={36},
  pages={13294--13307},
  year={2023}
}

@article{meng2024conditional,
  title={A conditional diffusion model with fast sampling strategy for remote sensing image super-resolution},
  author={Meng, Fanen and Chen, Yijun and Jing, Haoyu and Zhang, Laifu and Yan, Yiming and Ren, Yingchao and Wu, Sensen and Feng, Tian and Liu, Renyi and Du, Zhenhong},
  journal={IEEE Transactions on Geoscience and Remote Sensing},
  year={2024},
  publisher={IEEE}
}

@article{liu2022diffusion,
  title={Diffusion model with detail complement for super-resolution of remote sensing},
  author={Liu, Jinzhe and Yuan, Zhiqiang and Pan, Zhaoying and Fu, Yiqun and Liu, Li and Lu, Bin},
  journal={Remote Sensing},
  volume={14},
  number={19},
  pages={4834},
  year={2022},
  publisher={MDPI}
}

@inproceedings{yu2024scaling,
  title={Scaling up to excellence: Practicing model scaling for photo-realistic image restoration in the wild},
  author={Yu, Fanghua and Gu, Jinjin and Li, Zheyuan and Hu, Jinfan and Kong, Xiangtao and Wang, Xintao and He, Jingwen and Qiao, Yu and Dong, Chao},
  booktitle={Proceedings of the IEEE/CVF conference on computer vision and pattern recognition},
  pages={25669--25680},
  year={2024}
}

@inproceedings{lin2024diffbir,
  title={Diffbir: Toward blind image restoration with generative diffusion prior},
  author={Lin, Xinqi and He, Jingwen and Chen, Ziyan and Lyu, Zhaoyang and Dai, Bo and Yu, Fanghua and Qiao, Yu and Ouyang, Wanli and Dong, Chao},
  booktitle={European conference on computer vision},
  pages={430--448},
  year={2024},
  organization={Springer}
}

@article{li2025diffusion,
  title={Diffusion models for image restoration and enhancement: a comprehensive survey},
  author={Li, Xin and Ren, Yulin and Jin, Xin and Lan, Cuiling and Wang, Xingrui and Zeng, Wenjun and Wang, Xinchao and Chen, Zhibo},
  journal={International Journal of Computer Vision},
  pages={1--31},
  year={2025},
  publisher={Springer}
}

@article{ho2020denoising,
  title={Denoising diffusion probabilistic models},
  author={Ho, Jonathan and Jain, Ajay and Abbeel, Pieter},
  journal={Advances in neural information processing systems},
  volume={33},
  pages={6840--6851},
  year={2020}
}

@article{dhariwal2021diffusion,
  title={Diffusion models beat gans on image synthesis},
  author={Dhariwal, Prafulla and Nichol, Alexander},
  journal={Advances in neural information processing systems},
  volume={34},
  pages={8780--8794},
  year={2021}
}

@article{saharia2022image,
  title={Image super-resolution via iterative refinement},
  author={Saharia, Chitwan and Ho, Jonathan and Chan, William and Salimans, Tim and Fleet, David J and Norouzi, Mohammad},
  journal={IEEE transactions on pattern analysis and machine intelligence},
  volume={45},
  number={4},
  pages={4713--4726},
  year={2022},
  publisher={IEEE}
}

@article{li2022srdiff,
  title={Srdiff: Single image super-resolution with diffusion probabilistic models},
  author={Li, Haoying and Yang, Yifan and Chang, Meng and Chen, Shiqi and Feng, Huajun and Xu, Zhihai and Li, Qi and Chen, Yueting},
  journal={Neurocomputing},
  volume={479},
  pages={47--59},
  year={2022},
  publisher={Elsevier}
}

@inproceedings{gao2023implicit,
  title={Implicit diffusion models for continuous super-resolution},
  author={Gao, Sicheng and Liu, Xuhui and Zeng, Bohan and Xu, Sheng and Li, Yanjing and Luo, Xiaoyan and Liu, Jianzhuang and Zhen, Xiantong and Zhang, Baochang},
  booktitle={Proceedings of the IEEE/CVF conference on computer vision and pattern recognition},
  pages={10021--10030},
  year={2023}
}

@article{kingma2014adam,
  title={Adam: A method for stochastic optimization},
  author={Kingma, Diederik P and Ba, Jimmy},
  journal={arXiv preprint arXiv:1412.6980},
  year={2014}
}

@article{qi2025advancing,
  title={Advancing Image Super-resolution Techniques in Remote Sensing: A Comprehensive Survey},
  author={Qi, Yunliang and Lou, Meng and Liu, Yimin and Li, Lu and Yang, Zhen and Nie, Wen},
  journal={arXiv preprint arXiv:2505.23248},
  year={2025}
}

@inproceedings{wu2023hsr,
  title={HSR-Diff: Hyperspectral image super-resolution via conditional diffusion models},
  author={Wu, Chanyue and Wang, Dong and Bai, Yunpeng and Mao, Hanyu and Li, Ying and Shen, Qiang},
  booktitle={Proceedings of the IEEE/CVF International Conference on Computer Vision},
  pages={7083--7093},
  year={2023}
}

@article{weng2025efficient,
  title={Efficient High-Frequency Texture Recovery Diffusion Model for Remote Sensing Image Super-Resolution},
  author={Weng, Wu-Ding and Zheng, Chao-Wei and Su, Jian-Nan and Chen, Guang-Yong and Gan, Min},
  journal={IEEE Transactions on Instrumentation and Measurement},
  year={2025},
  publisher={IEEE}
}

@article{miao2025research,
  title={Research on Cross-Sensor Remote Sensing Image Super-Resolution Method Based on Diffusion Models},
  author={Miao, Ru and Yang, Kai and Zhou, Ke and Song, Jia and Fu, Shihao and Liu, Cong and Wang, Yuanxing},
  journal={IEEE Journal of Selected Topics in Applied Earth Observations and Remote Sensing},
  year={2025},
  publisher={IEEE}
}

@article{wang2024super,
  title={Super-resolution image reconstruction method between Sentinel-2 and Gaofen-2 based on cascaded generative adversarial networks},
  author={Wang, Xinyu and Ao, Zurui and Li, Runhao and Fu, Yingchun and Xue, Yufei and Ge, Yunxin},
  journal={Applied Sciences},
  volume={14},
  number={12},
  pages={5013},
  year={2024},
  publisher={MDPI}
}

@article{michel2025revisiting,
  title={Revisiting remote sensing cross-sensor Single Image Super-Resolution: the overlooked impact of geometric and radiometric distortion},
  author={Michel, Julien and Kalinicheva, Ekaterina and Inglada, Jordi},
  journal={IEEE Transactions on Geoscience and Remote Sensing},
  year={2025},
  publisher={IEEE}
}

@article{zhu2025unidb,
  title={UniDB: A Unified Diffusion Bridge Framework via Stochastic Optimal Control},
  author={Zhu, Kaizhen and Pan, Mokai and Ma, Yuexin and Fu, Yanwei and Yu, Jingyi and Wang, Jingya and Shi, Ye},
  journal={arXiv preprint arXiv:2502.05749},
  year={2025}
}

@article{lipman2022flow,
  title={Flow matching for generative modeling},
  author={Lipman, Yaron and Chen, Ricky TQ and Ben-Hamu, Heli and Nickel, Maximilian and Le, Matt},
  journal={arXiv preprint arXiv:2210.02747},
  year={2022}
}

@Misc{Planet2025API,
  author       = {Planet Labs PBC},
  organization = {Planet},
  title        = {Planet Application Program Interface: In Space for Life on Earth},
  year         = {2025},
  url          = "https://api.planet.com"
}

@Article{Drusch2012Sentinel2,
  author  = {Drusch, M. and Del Bello, U. and Ciolini, S. and et al.},
  title   = {Sentinel-2: {ESA}'s Optical High-Resolution Mission for {GMES} Operational Services},
  journal = {Remote Sensing of Environment},
  year    = {2012},
  volume  = {120},
  pages   = {25--36},
  doi     = {10.1016/j.rse.2011.11.026}
}

@inproceedings{kopec2025supresdiffgan,
  title={SupResDiffGAN: A New Approach for the Super-Resolution Task},
  author={Kope{\'c}, Dawid and Koz{\l}owski, Wojciech and Wizerkaniuk, Maciej and Krutul, Dawid and Koco{\'n}, Jan and Zi{\k{e}}ba, Maciej},
  booktitle={Proceedings of the International Conference on Computational Science (ICCS)},
  year={2025}
}

@inproceedings{ronneberger2015u,
  title={U-net: Convolutional networks for biomedical image segmentation},
  author={Ronneberger, Olaf and Fischer, Philipp and Brox, Thomas},
  booktitle={International Conference on Medical image computing and computer-assisted intervention},
  pages={234--241},
  year={2015},
  organization={Springer}
}

@misc{von2022diffusers,
  title={Diffusers: State-of-the-art diffusion models},
  author={Von Platen, Patrick and Patil, Suraj and Lozhkov, Anton and Cuenca, Pedro and Lambert, Nathan and Rasul, Kashif and Davaadorj, Mishig and Wolf, Thomas},
  year={2022}
}
\end{document}